\documentclass[onecolumn, 12pt, 3p]{article}
\usepackage{amssymb}
\usepackage{hyperref}
\usepackage{graphicx}

\usepackage{listings}
\usepackage{xcolor}
\usepackage{appendix}
\usepackage{titlesec}

\newcommand{\rev}[1]{\textcolor{black}{#1}}

\definecolor{codegreen}{rgb}{0,0.6,0}
\definecolor{codegray}{rgb}{0.5,0.5,0.5}
\definecolor{codepurple}{rgb}{0.58,0,0.82}
\definecolor{backcolour}{rgb}{0.95,0.95,0.92}

\lstdefinestyle{mystyle}{
    backgroundcolor=\color{backcolour},   
    commentstyle=\color{codegreen},
    keywordstyle=\color{magenta},
    numberstyle=\tiny\color{codegray},
    stringstyle=\color{codepurple},
    basicstyle=\ttfamily\scriptsize,
    breakatwhitespace=false,         
    breaklines=true,                 
    captionpos=b,                    
    keepspaces=true,
    language=Python,
    morekeywords={int8, uint8, uint16, uint32, uint64, float32},
    numbers=left,                    
    numbersep=5pt,                  
    showspaces=false,                
    showstringspaces=false,
    showtabs=false,                  
    tabsize=2
}

\graphicspath{ {./Images/} }
\begin{document}
%\renewcommand{\labelenumii}{\arabic{enumi}.\arabic{enumii}}

%\begin{frontmatter}
%--- INSTRUCTIONS TO BE DELETED OR COMMENTED BEFORE SUBMISSION 
%--- END OF INSTRUCTIONS TO BE DELETED OR COMMENTED BEFORE SUBMISSION 
 


\title{ ars548\_ros: An ARS 548 RDI radar driver for ROS}


\author{F. Fernández-Calatayud \and
L. Coto \and D. Alejo \and J. J. Carpio \and F. Caballero
 \and L. Merino}
%\address[la1]{Universidad Pablo de Olavide, Service Robotics Lab, Spain,      \\ (ffercal@upo.es, lcotele@upo.es, jjcarjim@upo.es, fcaballero@upo.es, lmercab@upo.es)}
%\address[la2]{Universidad de Sevilla, Service Robotics Lab, Spain, dalejo@us.es}
\maketitle
\begin{abstract}
The ARS 548 RDI Radar is a premium model of the fifth generation of 77 GHz long-range radar sensors with new RF antenna arrays, which offer digital \rev{beamforming}. This radar measures independently the distance, speed, and angle of objects without any reflectors in one measurement cycle based on Pulse Compression with New Frequency Modulation. Unfortunately, to the best of our knowledge, there are no open-source drivers available for Linux systems to enable users to analyze the data acquired by the sensor. In this paper, we present a driver that can interpret the data from the ARS 548 RDI sensor and make it available over the Robot Operating System versions 1 and 2 (ROS and ROS2). Thus, these data can be stored, represented, and analyzed using the powerful tools offered by ROS. Besides, our driver offers advanced object features provided by the sensor, such as relative estimated velocity and acceleration of each object, its orientation and angular velocity. We focus on the configuration of the sensor and the use of our driver including its  filtering and representation tools. Besides, we offer a video tutorial to help in its configuration process. Finally, a dataset acquired with this sensor and an Ouster OS1-32 LiDAR sensor, to have baseline measurements, is available so that the user can check the correctness of our driver.
 
keywords: ROS ; Linux driver ; Radar sensor ; ARS 548 

\end{abstract}

%\end{frontmatter}

%\linenumbers

\section*{Code metadata}
\begin{table}[h!]
\footnotesize
\begin{tabular}{p{0.46\textwidth}p{0.46\textwidth}}
\hline
Current code version &  1.0.\rev{1}\\

Permanent link to code/repository used for this code version &\url{https://github.com/robotics-upo/ars548_ros} \\

Permanent link to Reproducible Capsule & \url{}\\

Legal Code License   &  BSD 3-Clause License\\

Code versioning system used & Git \\

Software code languages, tools, and services used & \rev{C++} \\

Compilation requirements, operating environments \& dependencies & Linux\\

If available Link to developer documentation/manual &  \url{https://github.com/robotics-upo/ars548_ros/blob/master/Readme.md} \\

Support email for questions & \url{dalejo@us.es} \\
\hline
\end{tabular}
%\caption{Code metadata}
\label{codeMetadata} 
\end{table}
\textbf{Software metadata} \\
\begin{table}[h!]
\footnotesize
\begin{tabular}{{p{0.46\textwidth}p{0.46\textwidth}}}

\hline
Current software version & 1.0.\rev{1} \\

Permanent link to executables of this version  &  \url{} \\

Permanent link to Reproducible Capsule & \\

Legal Software License & BSD 3-Clause License \\

Computing platforms/Operating Systems &  Linux \\

Installation requirements & ROS2 Humble or ROS Noetic \\

Support email for questions & \url{dalejo@us.es} \\

\hline
\end{tabular}
%\caption{Software metadata}
\label{executableMetadata} 
\end{table}

\section{Motivation and significance}

Autonomous navigation in reduced visibility scenarios, i.e. scenarios with the presence of fog, rain, dust or smoke, is still a great challenge for robots, as the prevalent sensors used for obstacle detection in robotics, i.e. Light Detection and Ranging sensors (LiDAR), RGB and RGB-D cameras, can significantly reduce their precision or become useless when operating under these conditions \cite{booker_jfr_2007,li_icra_2021}. Also, sensors onboard a smartphone can be used in robotic applications thanks to applications such as GetSensorData \cite{GUTIERREZ2022101186}, but the performance of most onboard sensors would also drop in reduced visibility scenarios. Therefore, different sensors should be employed to allow autonomous robots to be deployed in reduced visibility scenarios. 

One alternative can be the use of thermal infrared cameras for odometry and localization \cite{khattak2020keyframe,saputra2020deeptio}\rev{, but the problem with cameras is the unknown scale ratio}. In this paper, we focus on RAdio Detection And Ranging sensors (radar), as aerosols do not affect them because their wavelength is larger than most aerosol particles. Thus, radar sensors are able to penetrate rain, dust and smoke. However, using a larger wavelength also has drawbacks, as the measurements obtained by a radar sensor have worse distance and angular accuracy and resolution when compared to LiDAR measurements \cite{Mielle2019}.

Localization and navigation in reduced-visibility conditions is an active research field in autonomous car navigation. In fact, the vast majority of reduced-visibility datasets include radar information \cite{oxford_radar, seeing_through_fog, hong_radar_slam_2020}. 
A common approach is to perform data fusion \cite{alejo_fusion} of information perceived from LiDAR and radar sensors to ensure that the objects are detected in all circumstances. Unfortunately, the rotating radar sensors used in these datasets (such as the Navtech Radar CTS350-X) are too expensive and heavy to be included in a cost-effective robotic solution. 

In the past years, advances in Frequency-Modulated Continuous Wave (FMCW) radar, which are based on the Doppler effect to detect surrounding objects, have allowed the development of new radar sensors for robotic applications. Not only are these sensors able to detect the distance of nearby objects to them, but they can also estimate their relative velocity by taking advantage of the Doppler effect. This velocity information can be used to improve the odometry estimation in harsh environments \cite{radar_odometry_quist_2016}, which is fundamental for the safe autonomous navigation of robots. Still, their use for 3D navigation is challenging due to the low resolution and narrow vertical Field of View (FoV) of inexpensive models, such as the IWR6843 from Texas Instruments. 

% Then introduce the sensor and indicate its characteristics comparing it to previous radar sensors

\begin{figure}[!t]
\centering
\includegraphics[width=0.6\textwidth]{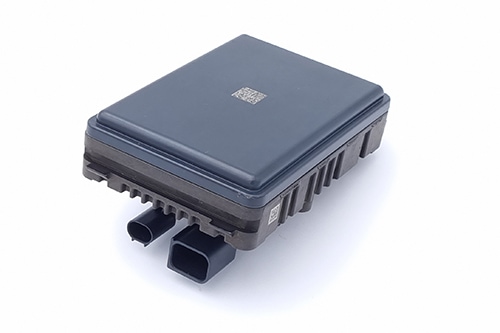}
\caption{Continental ARS 548 RDI high resolution radar sensor.}
\label{fig:Radar}
\end{figure}

 More recently, the ARS 548 RDI sensor from Continental (see Figure \ref{fig:Radar}) has been developed. It uses more advanced modulation techniques to better estimate the position and velocity of nearby objects. We believe that this sensor can be a good option for enabling a mobile robot or autonomous car to operate in reduced visibility scenarios \rev{ \cite{barzaghi, ars548_robust}. This radar }is a good compromise solution between inexpensive but feature-lacking FMCW radar sensors and 2D rotating radars used for automotive purposes. Unfortunately, we did not find any open-source driver that would allow us to acquire and process the data acquired from \rev{it.}%this sensor. 

%Finally another summary of the paper.
 
 In this paper, we describe the architecture of the \texttt{ars548\_ros} package. We highlight its interface to both versions of the Robotic Operating System \cite{ros2} (ROS), which are becoming the standard for robotic development at least for academic purposes, even though ROS2 is more industrially focused than its predecessor, ROS. To develop ROS and ROS2 interfaces is very convenient as it allows us to use their representation, logging, and processing tools. Hereinafter, we will use ROS when we refer to both versions of ROS. We illustrate the good performance of our driver by including data collected from real experiments with measurements of an OS1-32 LiDAR sensor from Ouster as a baseline. Finally, the driver is distributed under the BSD 3-Clause License in the hope that it will be helpful to the community.

\section{Software description}

\begin{figure}[!t]
\centering
\includegraphics[width=0.98\textwidth]{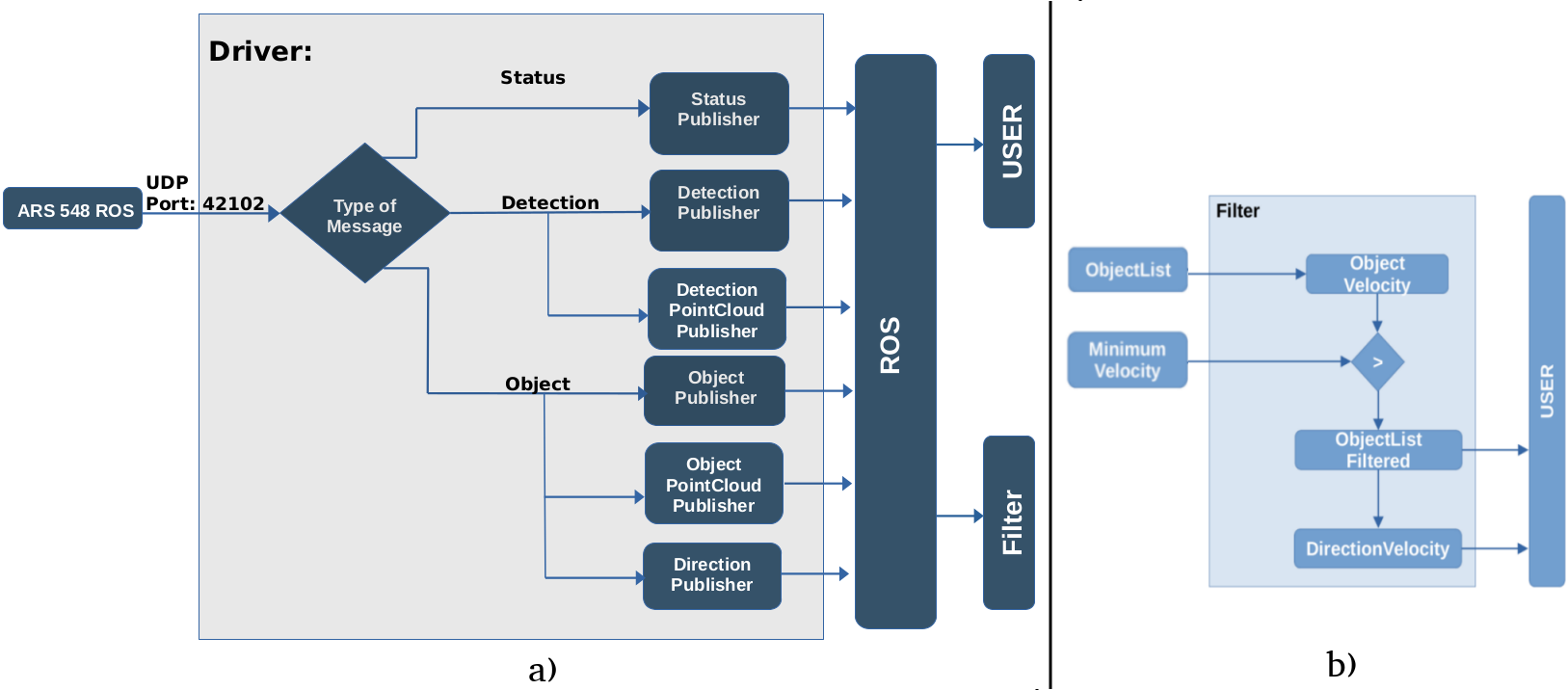}
\caption{a) Basic flow diagram of ARS 548 driver. b) Example application of the filter: detection of objects moving faster than a threshold velocity.}
\label{fig:Diagram}
\end{figure}

We developed our driver with two main objectives in mind. On the one hand, our driver generates standard ROS messages such as \texttt{PointCloud2} %\footnote{\url{https://docs.ros.org/en/humble/p/sensor_msgs/interfaces/msg/PointCloud2.html}} 
that include information on the position and relative velocity, if available, of the detected points and objects; and \texttt{PoseArray} that indicates the direction of the movement of the detected objects. By using standard messages, the user can take advantage of existing representation and processing tools available in ROS. On the other hand, we are interested in providing the user with all the information generated by the radar without any processing. To this end, we have created the following custom ROS messages: \texttt{ObjectList}, \texttt{DetectionList} and \texttt{Status} (see Figure \ref{fig:Diagram}a), which are a direct translation of the data structures sent by the sensor as described in the ARS 548 reference manual \cite{ars548_doc}. The messages have been included in Appendix \ref{appendix:messages} for the sake of completeness. Moreover, we can use the advanced information available in the Objects structure to filter them, so that only the required objects are taken into consideration (see Figure \ref{fig:Diagram}b and Section \ref{sec:filtering}). Finally, we include the definition of the \texttt{SensorConfiguration} message, which is used in the configuration tool that can change the parameters of the ARS 548 RDI radar (see Section \ref{sec:configuration}).

%This driver project consists on this number of folders and files: (You can see an scheme of this architecture on Figure \ref{fig:Architecture})
%\textit{The radar messages are classified in three types of message (SensorStatus, DetectionList and ObjectList). This messages often have a  list of objects or detections that they send so they have a lot of fields.}
%\textit{When our driver receives a message, it first classifies the type of message it has detected and starts transforming them into little endian (So when the user tries to read them, the values are the correct ones). When the transformation has ended, it stores the values in a structure that has the same fields as the message and finally it sends them to the user.}
%\textit{(A diagram of this functionality can be seen  on Figure \ref{fig:Diagram})}

\subsection{Software architecture}

Our solution consists of two ROS packages, with the following main functionalities: 

\begin{itemize}
    \item \textit{ars548\_driver}. This folder contains the libraries, code and functions needed for the actual driver to work. In particular, the \texttt{ars548\_driver} driver node receives the data from the radar via Ethernet link and transforms it into standard messages and custom messages (defined in \textit{ars548\_messages}), which are provided to the user via ROS middleware. On the other hand, \texttt{ars548\_filter} enables the user to distinguish different types of objects thanks to the advanced functionalities of the sensor. \rev{Finally, the \texttt{radar\_setup} utility allows the user to change some parameters of the radar, such as its maximum detection range, its IP address, and many more.}
    
    \item \textit{ars548\_messages}. In this folder we define the custom messages that are sent by our driver. These messages follow the data structure of the messages generated by the radar, as defined in \cite{ars548_doc}. 
\end{itemize}

%\begin{figure}[!t]
%\centering
%\includegraphics[width=0.5\textwidth]{Filter.png} %EstructuraDriver.png
%\caption{Example application of the filter: detection of objects moving faster than a threshold velocity.} 
%\label{fig:Architecture}
%\end{figure}

The full driver source code is publicly available in our GitHub repository \cite{repo}. \rev{P}lease refer to Appendix \ref{appendix:b} for the additional contents associated with the paper.

 \subsection{Software functionalities}

This section highlights the main functionalities provided by the ARS 548 driver.

 \subsubsection{Reception and translation of messages.}
    
    The proposed driver is able to establish a UDP connection with the sensor and translate the different information from the sensor to custom and standard ROS messages, making the necessary endian translations by means of templatized C++ functions. 

    With this functionality, we provide the user with all the information from ARS 548 as emitted by the sensors with our custom messages. Additionally, we provide standard messages, which are easier to visualize and process with standard ROS packages.
    
 \subsubsection{Visualization of the results.}
    
\begin{figure}[!t]
\centering
\includegraphics[width=0.65\textwidth]{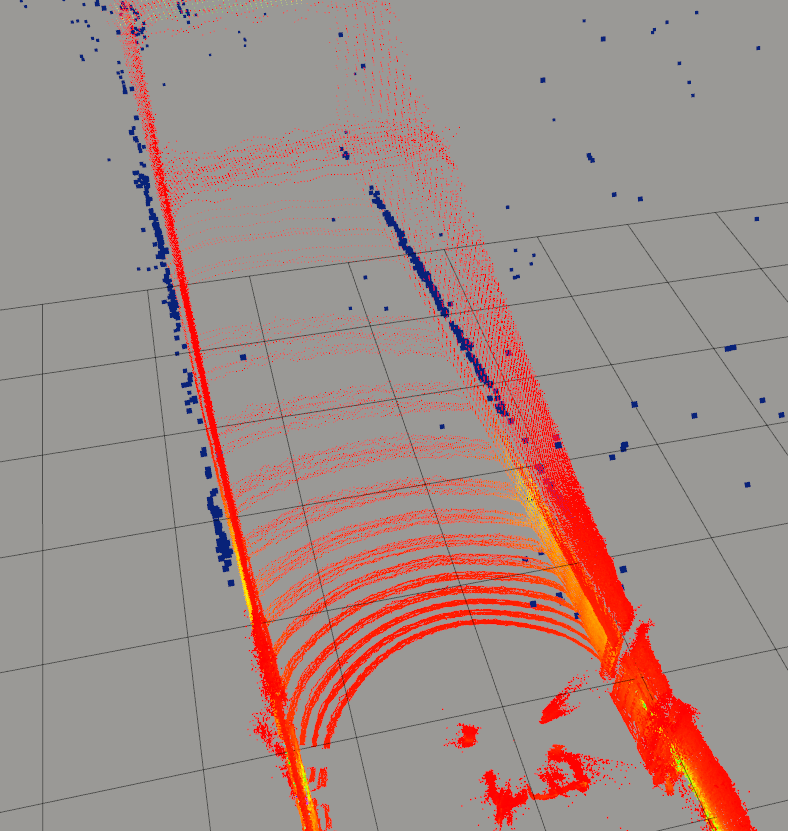}
\caption{Radar detection measurements captured by the ARS 548 RDI sensor in blue and OS1 \rev{LiDAR} measurements in red. Please note the alignment between LiDAR and radar measurements.}
\label{fig:Radar-lidar}
\end{figure}

    By generating the \texttt{PointCloud2} and \texttt{PoseArray} standard ROS messages, the user can easily represent the objects in the standard ROS visualizer (RViz \cite{rviz}). To this end, we provide the user with custom visualization configurations that can be loaded in RViz for easily representing the measurements from the radar sensor (see Figure \ref{fig:Diagram}a). You can see an example of the representation of the radar measurements in Figure \ref{fig:Radar-lidar}.

%\begin{figure}[!t]
%\centering
%\includegraphics[width=0.45\textwidth]{rviz.png}
%\caption{Radar detection measurements captured by the ARS 548 RDI sensor in blue and OS1 lidar measurements in red. The.}
%\label{fig:rviz}
%\end{figure}

 \subsubsection{Object filtering} \label{sec:filtering}

The \texttt{ars548\_ros} package provides the user with a filter able to interpret our custom messages and to provide filtered point clouds depending on the extended information generated by the sensor. The available information is defined in the \texttt{Object} message (see Appendix \ref{appendix:messages}). The filter can easily be  customized via inheritance just by overriding a method; please refer to our repository for more details. As an example, see Section \ref{sec:exp1} in which we used a filter to represent only moving vehicles in the environment using the speed estimation from the sensor.

\subsubsection{Time Stamp Options}

\rev{
We offer two different options to stamp the information received from the ARS 548:
}
\begin{itemize}

    \item \rev{\textbf{Keep the original stamp.} Use this option if the ARS 548 device is synchronized with the local PC, which can be achieved by means of the gPTP protocol. To this end, the user PC has to be configured in master mode, so that the sensor can detect it\footnote{Please refer to \url{https://linuxptp.sourceforge.net/} for configuring it.}. %If the machines are synchronized, we offer the option to keep the received stamp.% r This can be very convenient in high-speed applications, where a correct stamp can be vital.
    }
    \item \rev{\textbf{Override stamp with local time.} In case there is not synchronization between the radar sensor and the PC, for example if its Ethernet interface does not support it, we offer an option to override the received timestamp using the local reception time. %It can be useful when the RADAR is not synchronized properly or as a first step to check communications with the ARS 548 RDI. When the driver performs the stamp override, the user is able to easily represent the data properly in RViz and use it in their programs. 
    However, there could be a loss of precision due to incorrect stamping that can be noticeable in high-speed navigation. }
\end{itemize}

\subsubsection{Radar configuration}
\label{sec:configuration}

\rev{We have included a basic utility that allows the user to change the configuration parameters of the ARS 548. The configuration parameters of the device can be divided into the following categories. The reader is referred to the  \texttt{SenorConfiguration.msg} file in Appendix \ref{appendix:messages} to see the detailed list of the configuration parameters:
}

\begin{enumerate}
    \item \rev{\textbf{Mounting position.} This set of parameters allows the user to configure the mounting position of the device w.r.t. the front axle of the vehicle (lateral, longitudinal and vertical parameters). Besides, the orientation of the device (yaw and pitch angles) and the plug (left or right) can be specified. These parameters are used together with the vehicle parameters in the auto-alignment feature of the ARS 548.} 
    \item \rev{\textbf{Vehicle parameters.} They allow the user to specify the main parameters of the car, which are width, length,  height and wheelbase.
    \item \textbf{Radar configuration.} They change the mode of operation of the radar. You can configure its maximum detection distance (from 99 to 1500 meters), and choose between three different frequency slots. They can also refer to internal configuration of the radar, such as its IP address and power saving options, to name a few. }
    
\end{enumerate} 

\section{Illustrative examples}

\begin{figure*}[!t]
\centering
\includegraphics[width=\textwidth]{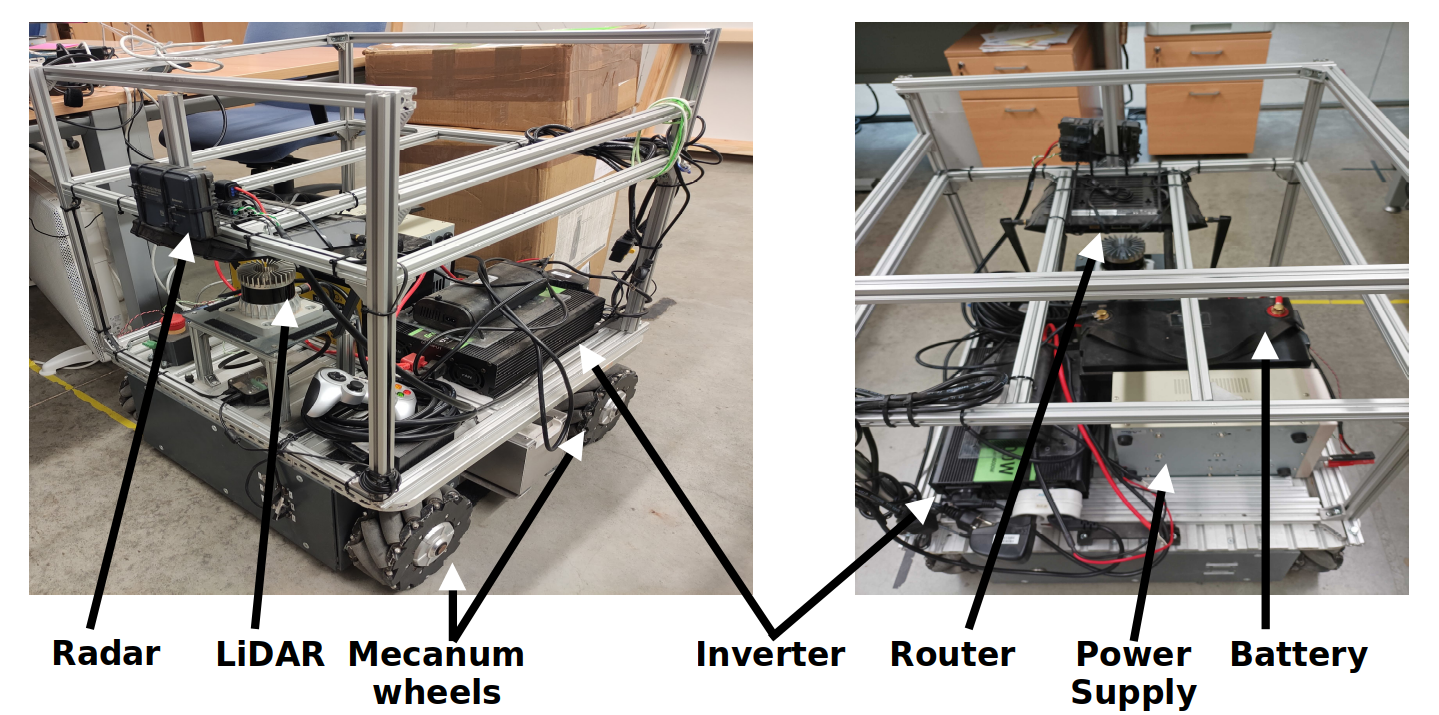}
\caption{Our ARCO robotic platform equipped with both an ARS 548 RDI sensor and an OS1-32 LiDAR sensor. The necessary equipment for the experiments is pointed out on the Figure.}
\label{fig:ARCO}
\end{figure*}

The driver presented in this paper has been experimentally tested in field experiments with the robotic ARCO platform \cite{arco} (see Figure \ref{fig:ARCO}). In particular, we describe here two example scenarios. First, a traffic monitoring scenario, in which the velocity estimation offered by the radar sensor can be of great use to distinguish the road traffic from the background. Second, a typical robotic scenario, in which the robot should be capable of both generating a map of the environment and localizing itself on it, in a problem that is usually referred to as Simultaneous Localization and Mapping (SLAM). 

\subsection{Experimental setup}

Our ARCO robotic platform has been equipped with an ARS 548 RDI sensor and an OS1-32 LiDAR sensor, which was used as a reference to confirm the validity of the measurements obtained by the radar sensor (see Figure \ref{fig:ARCO}). We used an MSI GF62 7RE Laptop with 32GB RAM \rev{and an 11th generation Intel Core i7 processor. The laptop is connected to} both sensors via Ethernet, with independent connections. For each experiment, we generated one or more ROSBag files with all the information from both sensors. In particular, with regards to ARS 548 RDI, we saved both our custom topics and the standard ones. The dataset is available at \cite{dataset}.

\subsection{Driver Performance}

\rev{We have performed a thirty-minute long experiment, in which we used our driver and visualized the data from the ARS 548 sensor in real time with RViz. Our goal is to check the usual load and memory usage using the Ubuntu built-in system monitor. During the experiment, the driver process used a mere 2\% of the processing power of one core. Furthermore, its memory usage was stable at around 7 MB. Therefore, the driver has a very low impact on the system resources.
}
%\textcolor{red}{TBC}

%\rev{
%Before performing any experiment, there are two steps that should be performed to get the best performance of the ARS 548.}

%\begin{enumerate}
    
 %   \item \rev{\textbf{Time synchronization. }}
  %  \item \rev{\textbf{Self calibration. } }
%\end{enumerate}

%\rev{
%According to the provided user manual, the }

\subsection{Experiment 1. Traffic monitoring}

\label{sec:exp1}

\begin{figure}[!t]
\centering
\includegraphics[width=0.8\textwidth]{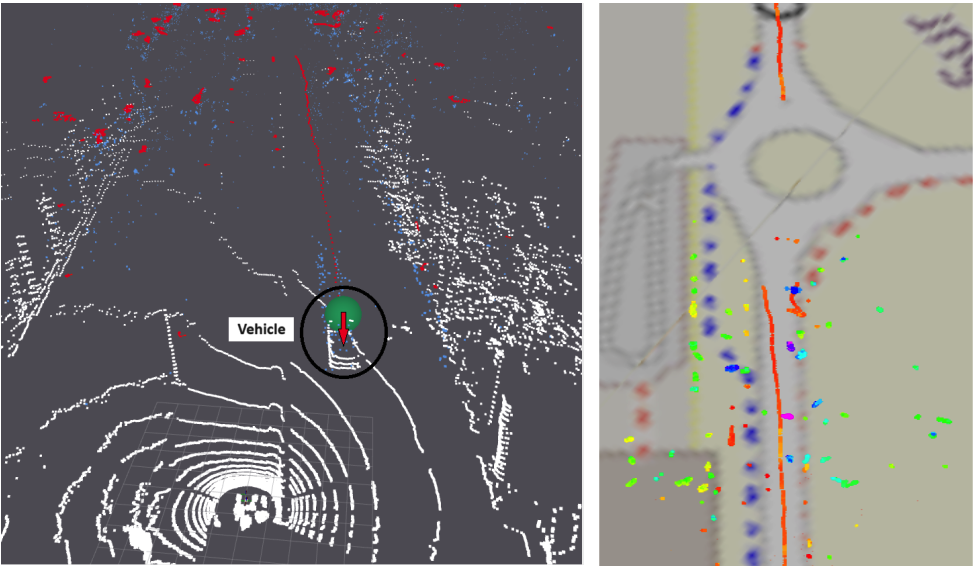}
\caption{Snapshot of Experiment 1. (left) Radar detections \rev{in} blue, radar objects \rev{in} red, radar object moving faster than 10 km/h \rev{in} a green sphere and LiDAR measurements \rev{in} white. (right) Geolocalized radar measurements represented over a map from OpenStreetMap \cite{OpenStreetMap}. Red lines represent cumulative radar detections of two different vehicles over the map of the environment. }
\label{fig:experiment1}
\end{figure}

In this experiment, our ARCO robotic platform remained static in the proximity of a road. The objective here is to use the ARS 548 RDI sensor to monitor the road traffic nearby, taking advantage of its medium-long range (up to 300 m) and its capability of measuring the speed of the detected objects. We use the object filtering feature of our driver (see Figure \ref{fig:Diagram}b) obtaining an object point cloud that contains vehicles moving faster than 10 km/h, which are assumed to be road traffic (see Figure \ref{fig:experiment1}). 

\subsection{Experiment 2. SLAM}
In the second experiment, our ARCO robotic platform navigates within an open environment while estimating its trajectory through the information received by the radar. Figure \ref{fig:Radar-lidar} illustrates the alignment between the measurements from our driver and the LiDAR, verifying its proper functioning.

\begin{figure}[!t]
\centering
\includegraphics[width=0.7\textwidth]{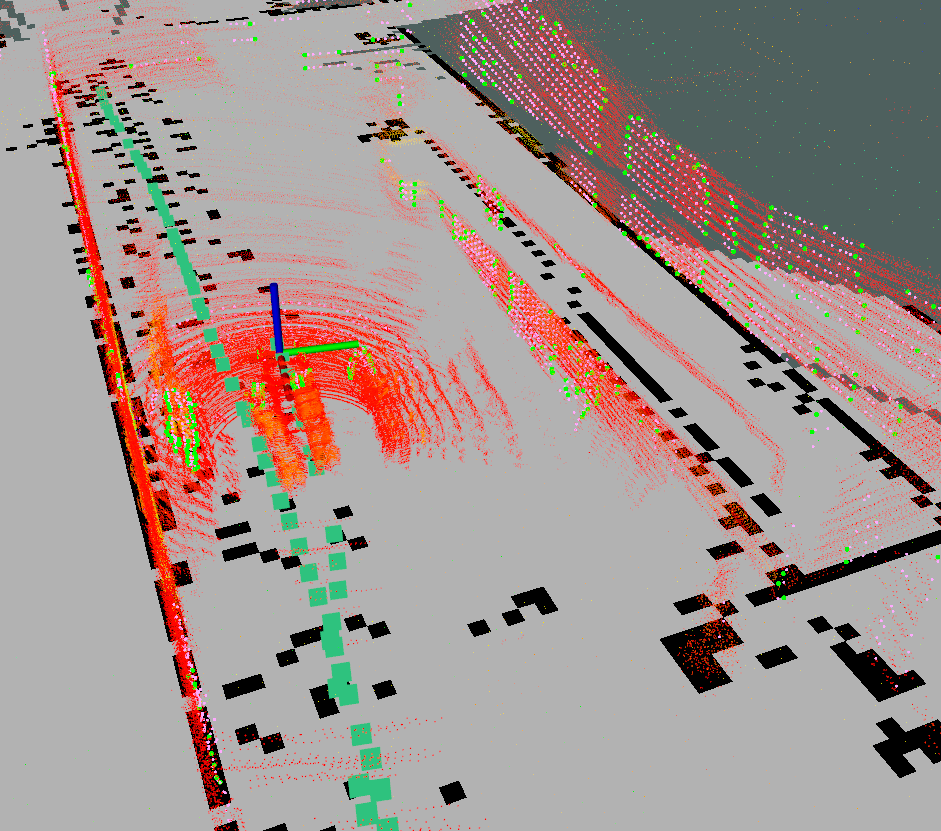}
\caption{The baseline trajectory generated by the LeGO-LOAM package is represented in large green dots over the generated map projection in 2D. The LiDAR measurements are represented in red dots, with small green dots representing features detected by Lego-LOAM.}
\label{fig:LeGO-LOAM}
\end{figure}

To validate the results of this experiment, we have employed the odometry generated by the LeGO-LOAM package \cite{legoloam2018} as a baseline, which relies on the LiDAR sensor for data acquisition (see Figure \ref{fig:LeGO-LOAM}). The data acquired from both the radar and LiDAR sensors will be used for future odometry and SLAM techniques, ensuring accurate mapping and localization, especially in hazardous scenarios where radar is employed.

\section{Impact}

This open source driver has been published under the BSD 3-Clause License in the hope of minimizing the development efforts of all users of the ARS 548 RDI sensor who would like to implement ROS-based applications with it. By doing so, the users can easily set up the driver and acquire its data with ROS interfaces and then focus on processing the data as needed for their own experiments, significantly speeding up their research and development process. \rev{The code is published in the GitHub platform \cite{repo}. }

Our main goal is to use this device for inspection robotics application carried out in reduced visibility environments, mainly due to weather phenomena such as rain, fog and snow or emergency situations involving smoke. These applications may include the inspection of different facilities such as offshore oil\&gas extraction plants, sewers and transmission and electric power lines \cite{rs16020355}. By using this sensor, we can ensure that inspection operations can be performed safely under all circumstances.

Another application field is autonomous driving. In fact, this is the main area of application for the ARS 548 RDI devices. The idea here is to incorporate this sensor into an autonomous vehicle to provide it with detailed information of surrounding objects including the type of object or vehicle and its relative velocity and acceleration. With this information, the user can design collision detection and resolution algorithms that guarantee the collision-free operation of the vehicle in spite of harsh weather or poor illumination circumstances.

\section{Conclusions}
This paper described a ROS driver to configure and interface with the ARS 548 RDI radar sensor. The driver works on both ROS and ROS2 versions, maximizing its impact on the community. The driver has been tested in two different field experiments, showcasing the advanced information from the sensor and the filtering capabilities of our driver. 

Future work \rev{will focus on implementing interfaces to allow the ROS modules from the user to provide the ARS 548 radar with information, such as estimated velocity and accelerations, to perform its provided auto-alignment function, which is adequate for automotive purposes. Finally, we are developing new radar odometry approaches, and LiDAR-radar data fusion mechanisms. }

\section*{Declaration of competing interest}
The authors declare that they have no known competing financial interests or personal relationships that could have appeared
to influence the work reported in this paper.

\section*{Data availability}

All the experimental data presented in this paper are openly available at \cite{dataset}.  

\section*{Acknowledgements}
\label{sec:ack0}
Authors would like to thank Fernando Amodeo for his invaluable help.
This work has been supported by the grants INSERTION PID2021-127648OB-C31 and RATEC PDC2022-133643-C21. Both grants
have been funded by the Spanish Ministry of Science, Innovation and Universities with grant number 10.13039/501100011033 and “European Union NextGenerationEU/PRTR”.

\appendix

\section{Custom ARS 548 messages}
\label{appendix:messages}

Below you can find the definition of the custom ROS messages defined in the \texttt{ars548\_messages} ROS package.

\subsection*{Detection.msg}
\begin{lstlisting}
float32 f_azimuthangle #Unaligned Detection Azimuth Angle
float32 f_azimuthanglestd #Azimuth Angle Std
uint8 u_invalidflags #Detection Invalid Flags 
float32 f_elevationangle #Unaligned Detection Elevation Angle
float32 f_elevationanglestd #Elevation Angle Std
float32 f_range #Detection Radial Distance
float32 f_rangestd #Radial Distance Std
float32 f_rangerate #Detection Radial Velocity
float32 f_rangeratestd #Radial Velocity Std
int8 s_rcs #Detection RCS
uint16 u_measurementid #Detection ID
uint8 u_positivepredictivevalue #Existence Probability
uint8 u_classification #Detection Classification
uint8 u_multitargetprobabilitym #Multi-Target Probability
uint16 u_objectid #Associated Object ID
uint8 u_ambiguityflag #Probability for resolved velocity ambiguity
uint16 u_sortindex #tbd
\end{lstlisting}
\subsection*{DetectionList.msg}
\begin{lstlisting}
std_msgs/Header header
uint64 crc #Checksum (E2E Profile 7) (Reserved)
uint32 length #Len (E2E Profile 7) (Reserved)
uint32 sqc #SQC (E2E Profile 7) (Reserved)
uint32 dataid #Data ID (E2E Profile 7) (Reserved)
uint32 timestamp_nanoseconds #Timestamp Nanoseconds
uint32 timestamp_seconds #Timestamp Seconds
uint8 timestamp_syncstatus #Timestamp Sync Status
uint32 eventdataqualifier #Event Data Qualifier (unused)
uint8 extendedqualifier #Extended Qualifier (unused)
uint16 origin_invalidflags #Sensor Position Invalid flags (unused)
float32 origin_xpos #Sensor X Position with reference to rear axle
float32 origin_xstd #Sensor X Position STD (unused)
float32 origin_ypos #Sensor Y Position with reference to rear axle
float32 origin_ystd #Sensor Y Position STD (unused)
float32 origin_zpos #Sensor Z Position with reference to rear axle
float32 origin_zstd #Sensor Z Position STD (unused)
float32 origin_roll #Sensor Roll Angle (unused)
float32 origin_rollstd #Sensor Roll Angle STD (unused)
float32 origin_pitch #Sensor Pitch Angle with alignment correction
float32 origin_pitchstd #Sensor Pitch Angle STD
float32 origin_yaw #Sensor Yaw Angle with alignment correction
float32 origin_yawstd #Sensor Yaw Angle STD
uint8 list_invalidflags #Invalid flags (unused)
Detection[800] list_detections #Detection Array
float32 list_radveldomain_min #Ambiguity free Doppler velocity range Min
float32 list_radveldomain_max #Ambiguity free Doppler velocity range Max
uint32 list_numofdetections #Number of Detections
float32 aln_azimuthcorrection #Azimuth Alignment Correction
float32 aln_elevationcorrection #Elevation Alignment Correction
uint8 aln_status #Status of alignment
\end{lstlisting}
\subsection*{Object.msg}
\begin{lstlisting}
uint16 u_statussensor #tbd
uint32 u_id #Unique ID of object
uint16 u_age #Age of object
uint8 u_statusmeasurement #Object Status
uint8 u_statusmovement #Object Movement Status
uint16 u_position_invalidflags #tbd
uint8 u_position_reference #Reference point position
float32 u_position_x #X Position of reference point
float32 u_position_x_std #X Position Std
float32 u_position_y #Y Position of reference point
float32 u_position_y_std #Y Position Std
float32 u_position_z #Z Position of reference point
float32 u_position_z_std #Z Position Std
float32 u_position_covariancexy #Covariance X Y
float32 u_position_orientation #Object Orientation
float32 u_position_orientation_std #Orientation Std
uint8 u_existence_invalidflags #unused
float32 u_existence_probability #Probability of Existence
float32 u_existence_ppv #unused
uint8 u_classification_car #Car Classification Probability
uint8 u_classification_truck #Truck Classification Probability
uint8 u_classification_motorcycle #Motorcycle Classification Probability
uint8 u_classification_bicycle #Bicycle Classification Probability
uint8 u_classification_pedestrian #Pedestrian Classification Probability
uint8 u_classification_animal #Animal Classification Probability
uint8 u_classification_hazard #Hazard Classification Probability
uint8 u_classification_unknown #Unknown Classification Probability
uint8 u_classification_overdrivable #unused
uint8 u_classification_underdrivable #unused
uint8 u_dynamics_absvel_invalidflags #unused
float32 f_dynamics_absvel_x #X Absolute Velocity
float32 f_dynamics_absvel_x_std #X Absolute Velocity Std
float32 f_dynamics_absvel_y #Y Absolute Velocity
float32 f_dynamics_absvel_y_std #Y Absolute Velocity Std
float32 f_dynamics_absvel_covariancexy #Covariance Absolute Velocity X Y
uint8 u_dynamics_relvel_invalidflags #unused
float32 f_dynamics_relvel_x #X Relative Velocity
float32 f_dynamics_relvel_x_std #X Relative Velocity Std
float32 f_dynamics_relvel_y #Y Relative Velocity
float32 f_dynamics_relvel_y_std #Y Relative Velocity Std
float32 f_dynamics_relvel_covariancexy #Covariance Relative Velocity X Y
uint8 u_dynamics_absaccel_invalidflags #unused
float32 f_dynamics_absaccel_x #X Absolute Acceleration
float32 f_dynamics_absaccel_x_std #X Absolute Acceleration Std
float32 f_dynamics_absaccel_y #Y Absolute Acceleration
float32 f_dynamics_absaccel_y_std #Y Absolute Acceleration Std
float32 f_dynamics_absaccel_covariancexy #Covariance Absolute Acceleration X Y
uint8 u_dynamics_relaccel_invalidflags #unused
float32 f_dynamics_relaccel_x #X Relative Acceleration
float32 f_dynamics_relaccel_x_std #X Relative Acceleration Std
float32 f_dynamics_relaccel_y #Y Relative Acceleration
float32 f_dynamics_relaccel_y_std #Y Relative Acceleration Std
float32 f_dynamics_relaccel_covariancexy #Covariance Relative Acceleration X Y
uint8 u_dynamics_orientation_invalidflags #unused
float32 u_dynamics_orientation_rate_mean #Object Orientation Rate
float32 u_dynamics_orientation_rate_std #Orientation Rate Std
uint32 u_shape_length_status #(unused) Shape Length Status
uint8 u_shape_length_edge_invalidflags #(unused) Invalid Flags Shape Length 
float32 u_shape_length_edge_mean #Mean Shape Length
float32 u_shape_length_edge_std #(unused) Shape Length Std
uint32 u_shape_width_status #(unused) Shape Width Status 
uint8 u_shape_width_edge_invalidflags #(unused) Invalid Flags Shape Width
float32 u_shape_width_edge_mean #Mean Shape Width
float32 u_shape_width_edge_std #(unused) Shape Width Std 
\end{lstlisting}
\subsection*{ObjectList.msg}
\begin{lstlisting}
std_msgs/Header header
uint64 crc #Checksum (E2E Profile 7) (Reserved)
uint32 length #Len (E2E Profile 7) (Reserved)
uint32 sqc #SQC (E2E Profile 7) (Reserved)
uint32 dataid #Data ID (E2E Profile 7) (Reserved)
uint32 timestamp_nanoseconds #Timestamp Nanoseconds
uint32 timestamp_seconds #Timestamp Seconds
uint8 timestamp_syncstatus #Timestamp Sync Status
uint32 eventdataqualifier #(unused) Event Data Qualifier
uint8 extendedqualifier #(unused) Extended Qualifier 
uint8 objectlist_numofobjects #Number of Objects
Object[50] objectlist_objects #Object Array
\end{lstlisting}
\subsection*{Status.msg}
\begin{lstlisting}
uint32 timestamp_nanoseconds #Timestamp Nanoseconds
uint32 timestamp_seconds #Timestamp Seconds
uint8 timestamp_syncstatus #Timestamp Sync Status
uint8 swversion_major #Software version (major)
uint8 swversion_minor #Software version (minor)
uint8 swversion_patch #Software version (patch)
float32 longitudinal #Longitudinal sensor position (AUTOSAR)
float32 lateral #Lateral sensor position (AUTOSAR)
float32 vertical #Vertical sensor position (AUTOSAR)
float32 yaw #Sensor yaw angle (AUTOSAR)
float32 pitch #Sensor pitch angle (AUTOSAR)
uint8 plugorientation #Orientation of plug
float32 length #Vehicle length
float32 width #Vehicle width
float32 height #Vehicle height
float32 wheelbase #Vehicle wheelbase
uint16 maximundistance #Maximum detection distance
uint8 frequencyslot #Center frequency
uint8 cycletime #Cycle time
uint8 timeslot #Cycle offset
uint8 hcc #Country code
uint8 powersave_standstill #Power saving in standstill
uint32 sensoripaddress_0 #Sensor IP address
uint32 sensoripaddress_1 #Reserved
uint8 configurationcounter #Counter that counts up if new configuration has been received and accepted
uint8 status_longitudinalvelocity #Signals if current VDY is OK or timed out
uint8 status_longitudinalacceleration #Signals if current VDY is OK or timed out
uint8 status_lateralacceleration #Signals if current VDY is OK or timed out
uint8 status_yawrate #Signals if current VDY is OK or timed out
uint8 status_steeringangle #Signals if current VDY is OK or timed out
uint8 status_drivingdirection #Signals if current VDY is OK or timed out
uint8 status_characteristicspeed #(Unused) Signals if current VDY is OK or timed out.
uint8 status_radarstatus #Signals if Radar Status is OK
uint8 status_voltagestatus # Bitfield to report under- and overvoltage errors
uint8 status_temperaturestatus #Bitfield to report under- and overtemperature errors
uint8 status_blockagestatus #Current blockage state and blockage self test state.
\end{lstlisting}
\subsection*{SensorConfiguration.msg}
\begin{lstlisting}
float32 longitudinal #Longitudinal sensor position (AUTOSAR)
float32 lateral #Lateral sensor position (AUTOSAR)
float32 vertical #Vertical sensor position (AUTOSAR)
float32 yaw #Sensor yaw angle (AUTOSAR)
float32 pitch #Sensor pitch angle (AUTOSAR)
uint8 plugorientation #Orientation of plug
float32 length #Vehicle length
float32 width #Vehicle width
float32 height #Vehicle height
float32 wheelbase #Vehicle wheelbase
uint16 maximumdistance #Maximum detection distance
uint8 frequencslot #Center frequency (if MaximumDistance < 190 m only Mid can be selected)
uint8 cycletime #Cycle time
uint8 timeslot #Cycle offset
uint8 hcc #Country code
uint8 powersave_standstill #Power saving in standstill
uint32 sensoripaddress_0 #Sensor IP address
uint32 sensoripaddress_1 #Reserved (Configure as 2852025457 or 169.254.116.113)
uint8 newsensormounting #Flag if new sensor mounting position shall be configured
uint8 newvehicleparameters #Flag if new vehicle parameters position shall be configured
uint8 newradarparameters #Flag if new radar parameter shall be configured
uint8 newnetworkconfiguration # Flag if new IP address shall be configured
\end{lstlisting}

\section{Supplementary data}
\label{appendix:b}

The video tutorial for configuring the sensor and the driver can be found at \cite{tutorial}. The repository of the driver can be found at \cite{repo} and the acquired dataset can be found at \cite{dataset}.

\bibliographystyle{elsarticle-num} 
%\bibliography{radar_bib}

%\listoffigures

%\textit{If the software repository you used supplied a DOI or another
%Persistent IDentifier (PID), please add a reference for your software
%here. For more guidance on software citation, please see our guide for
%authors or \href{https://f1000research.com/articles/9-1257/v2}{this
%  article on the essentials of software citation by FORCE 11}, of
%which Elsevier is a member.}

%\large{\textbf{Reminder: Before you submit, please delete all 
%the instructions in this document, 
%including this paragraph. 
%Thank you!}}

\end{document}